\documentclass{article}

\usepackage[preprint, nonatbib]{nips_2018}
\usepackage[english]{babel}
\usepackage[utf8x]{inputenc}
\usepackage[T1]{fontenc}
\usepackage[justification=centering]{caption}
\usepackage{algorithm}
\usepackage{algpseudocode}
\usepackage{amsmath}
\usepackage{graphicx}
\usepackage{subcaption}
\usepackage[colorinlistoftodos]{todonotes}
\usepackage[colorlinks=true, allcolors=blue]{hyperref}
\usepackage{csquotes}
\usepackage{lipsum}

\begin{document}
\title{Positional Cartesian Genetic Programming}
\author{DG Wilson, Julian F. Miller, Sylvain Cussat-Blanc, Hervé Luga}

\maketitle

\begin{abstract}
Cartesian Genetic Programming (CGP) has many modifications across a variety of
implementations, such as recursive connections and node weights. Alternative
genetic operators have also been proposed for CGP, but have not been fully
studied. In this work, we present a new form of genetic programming based on a
floating point representation. In this new form of CGP, called Positional CGP,
node positions are evolved. This allows for the evaluation of many different
genetic operators while allowing for previous CGP improvements like recurrency.
Using nine benchmark problems from three different classes, we evaluate the
optimal parameters for CGP and PCGP, including novel genetic operators.
\end{abstract}

\section{Introduction}

Cartesian Genetic Programming (CGP) is a form of Genetic Programming (GP) where
program components are represented as functional nodes in on two dimensional
grid. \cite{Miller2011} Connections between these nodes are made based on their
Cartesian coordinates and create a final computational structure. The node
coordinates, the function of each node, and occasionally function parameters or
node weights are encoded in a genome which is evolved using a $1+\lambda$ EA.

Originally created for circuit design, CGP has since been shown to have
impressive results in image processing \cite{Harding2013}, creating neural
networks \cite{Khan2011} \cite{miller2017developmental}, and playing video games
\cite{wilson2018evolving}. It can create understandable computational structures
which illuminate solutions to the computational problems used as evolutionary
fitness metrics. For example, in \cite{wilson2018evolving}, simple programs with
constant or oscillatory behavior were generated by CGP, demonstrating effective
strategies for certain video games. Despite their simplicity, many were
competitive with or better than state of the art artificial agents.

CGP is an instance of graph-based GP, an attractive representation for
computational structures given that they can reuse subgraph components and are
used in many areas of computer science and engineering. In this work, we use
ideas from other forms of graph-based GP to design new mutation and crossover
methods for CGP. We also examine improvements to CGP that have been proposed,
evaluating them as hyper-parameters and using a parameter search to determine
when they are effective. These genetic operators and CGP enhancements are all
evaluated as hyper-parameters on nine different benchmark problems, with three
problems from each of the domains of classification, regression, and
reinforcement learning.

Some of the new genetic operators are made possible by using a floating point
representation of the CGP genome, as is done in \cite{Clegg2007}, with the
addition of 'snapping' connections which form connections to their nearest
target node. Beyond enabling certain genetic operators, this representation
allows for the evolution of the node positions itself, adding a new dimension to
the CGP evolution. We term this representation Positional Cartesian Genetic
Programming (PCGP) and evaluate it using the same hyper-parameter search used to
evaluate CGP.

\section{Cartesian Genetic Programming}

In its original formulation, CGP nodes are arranged in a rectangular grid of $R$
rows and $C$ columns \cite{Miller2000}. Nodes are allowed to connect to any node
from previous columns based on a connectivity parameter $L$ which sets the
number of columns back a node can connect to; for example, if $L=1$, nodes could
connect to the previous column only. In this paper, as in others, $R=1$, meaning
that all nodes are arranged in a single row.

\begin{figure*}[h]
  \centering
  \includegraphics[width=0.8\textwidth]{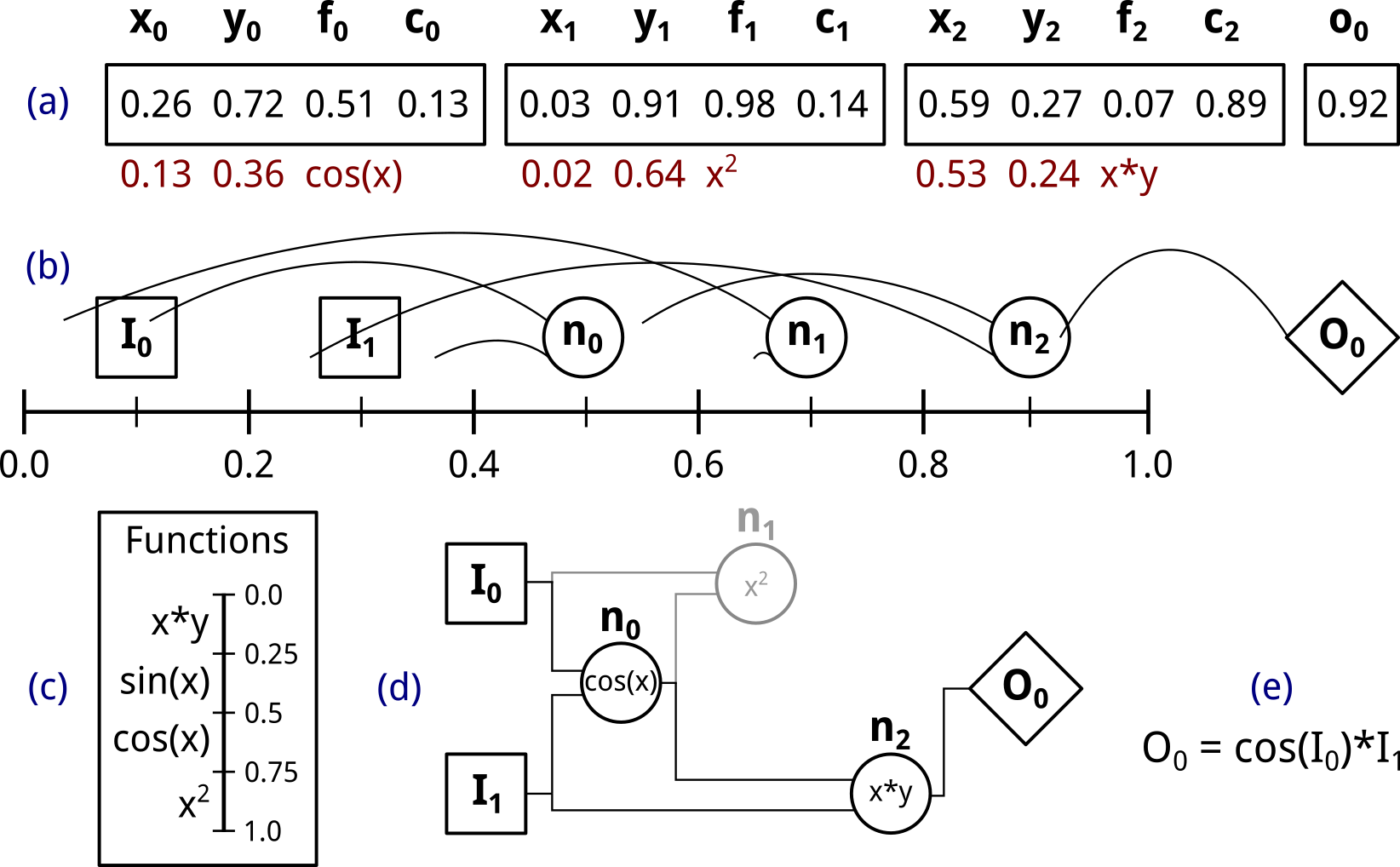}
  \caption{Translation of a floating point CGP genome into a program. The genome
    (a) is converted into positions and functions. The positions are informed by
    multiplying the connection genes, $x_n$ and $y_n$, by the position of the
    nodes, shown in (b). The functions are found by indexing the function gene
    $f_n$ in the Functions table (c). The resultant graph (d) is formed by
    ``snapping'' each connection to the closest node. As no output path uses
    $n_1$, this node is inactive or ``junk''. Finally, the graph can be read as
    a simple program (e).}
  \label{fig:fp_cgp}
\end{figure*}

In this work, a floating point representation of CGP is used. A similar
representation was previously used in \cite{Clegg2007}, but involved translation
from the traditional integer CGP representation to floating point. Here, floats
are used throughout. All genes are floating point numbers in $[0.0, 1.0]$, which
correspond to the connections of each node $n$, $x_n$ and $y_n$, the node
function $f_n$, and a parameter gene $c_n$ which can be used for node weights or
as a part of the function. Nodes are evenly spaced in one dimension between 0
and 1, with equal space around each node. Connections are formed by converting
the connection genes $x_n$ and $y_n$ to coordinates by multiplying the genes by
the node position, and then ``snapping'' these branches to the nearest node. An
example of this process is shown in \autoref{fig:fp_cgp}.

Each program output has a corresponding gene which connects to a node in the
graph. The output gene $o_n$ specifies a connection which then ``snaps'' to the
nearest node. By following connections back from the program outputs, an output
program graph can be constructed. In practice, only a small portion of the nodes
described by a CGP chromosome will be connected to its output program graph.
These nodes which are used are called ``active'' nodes here, whereas nodes that
are not connected to the output program graph are referred to as ``inactive'' or
``junk'' nodes. While these nodes do not actively contribute to the program's
output, they have been shown to aid evolutionary search \cite{Miller2006},
\cite{Vassilev2000}, \cite{Yu2001}.

\begin{figure*}[h]
  \centering
  \includegraphics[width=0.8\textwidth]{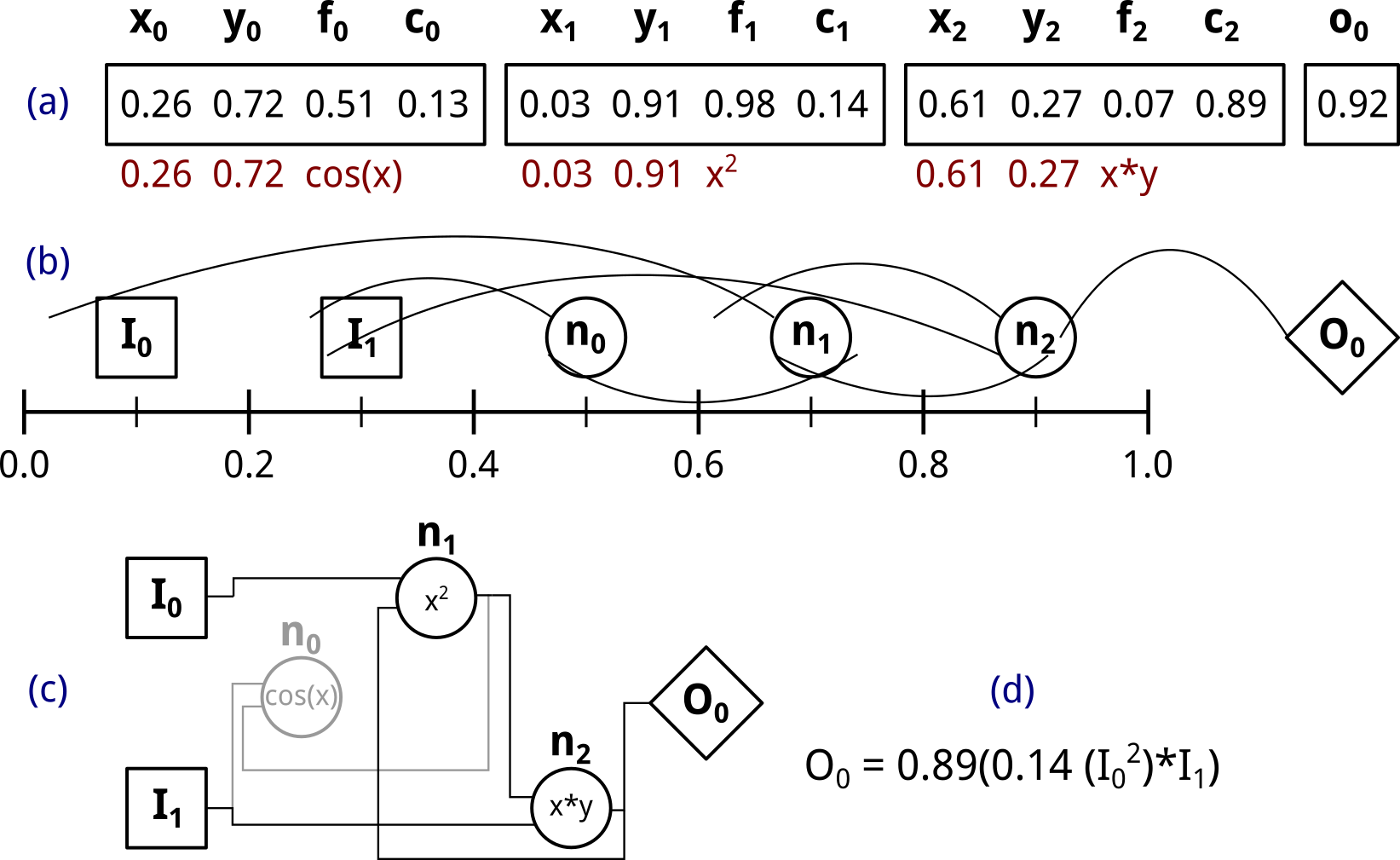}
  \caption{
    The same genome (a) as in \autoref{fig:fp_cgp}, but using a recurrency of
    $r=1.0$ and node weights $w=1$. The recurrency parameter changes the
    connection position calculation and allows nodes to connect to downstream
    nodes on their right (b). The output program graph (c) can then have
    recursive connections. Here, node weights are also used, modifying the final
    program (d).}
  \label{fig:fp_rcgp}
\end{figure*}

Two established CGP modifications are explored in this work: recurrent CGP and
node weights. In recurrent CGP \cite{Turner2014} (RCGP), a recurrency parameter
was introduced to express the likelihood of creating a recurrent connection;
when $r=0$, standard CGP connections were maintained, but $r$ could be increased
by the user to create recurrent programs. This work uses a slight modification
of the meaning of $r$, but the idea remains the same. Here, the final connection
position is modified by r:

\begin{equation}
  p_{x_n} = x_n(r(1.0 - p_n) + p_n)
\end{equation}

where $p_n$ is the position of node $n$, $x_n$ is its connection gene, and
$p_{x_n}$ is the position of the final connection. When $r=0.0$, this is as in
standard floating point CGP, as presented in \autoref{fig:fp_cgp}, where
$p_{x_n} = x_np_n$. when $r=1.0$, the connection positions are simply the gene
values, $p_{x_n} = x_n$. An example of this is shown in \autoref{fig:fp_rcgp}.
$r$ in this work therefore indicates the end of the possible range of
connections for a node, from that node's position $p_n$ to the end of the
positional space, 1.0.

In \autoref{fig:fp_rcgp}, node weights are also used. In this scheme, the output
of each node is multiplied by its parameter gene $c_n$. This CGP modification
has allowed for differentiable CGP \cite{izzo2017differentiable} and is referred
to in this work by the binary hyper-parameter $w$, which is true ($w=1$) when
weights are used.

\section{Positional Cartesian Genetic Programming}

With floating point CGP as a base, Positional CGP introduces a small
modification which allows for many possibilities. Each node also has a position
gene, $p_n$, which determines the position of the node, instead of spacing each
node equally between 0.0 and 1.0. In CGP, a connection has equal probability of
connecting to each node previous to its parent. In PCGP, this probability is
evolved based on the positions of each node.

Evolving the node positions complicates the role of the inputs, however. In
SM-CGP, where it also isn't certain the graph will include inputs, program input
is a function which nodes can choose \cite{Harding2010}. In this work, we chose
to place input in an evolved space to the left of the linear node space,
ensuring that nodes form connections to inputs while allowing the inputs to also
form their own connection distributions through evolution. Each inputs has a
positional gene, $i_n$, which is multiplied by a hyper-parameter which
determines the start of the input space, $I_{start}$. This input position gene
is initialized like all other genes, randomly in a uniform distribution in [0.0,
  1.0]. Node position calculation is modified with $I_{start}$ to contain the
entire space, including the input space:

\begin{equation}
  p_{x_n} = x_n((r(1.0 - p_n) + p_n) - I_{start}) + I_{start}
\end{equation}

\begin{figure*}[h]
  \centering
  \includegraphics[width=0.8\textwidth]{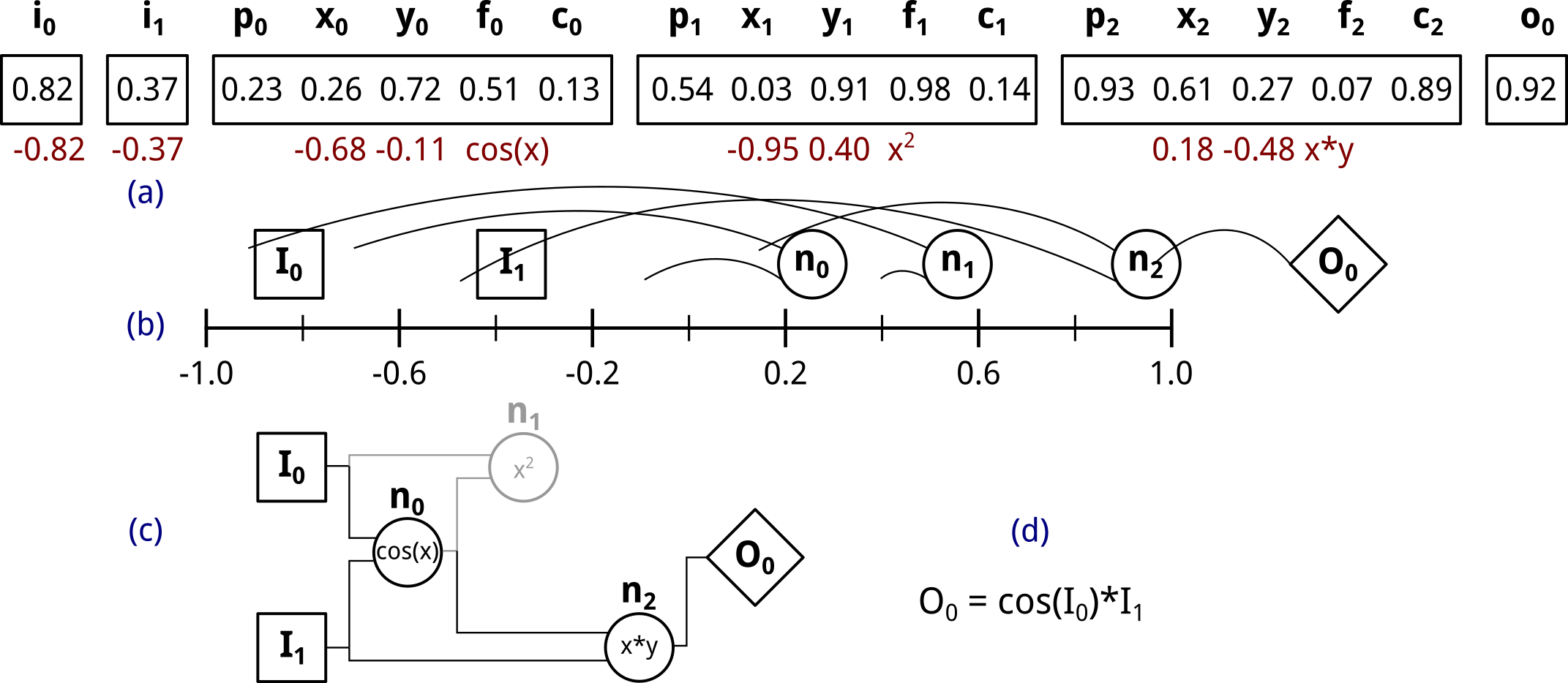}
  \caption{A PCGP genome (a), including input $i_n$ and positional $p_n$ genes.
    These are translated to input and node positions (b) and connection
    positions ``snap'' to the nearest node, as in floating point CGP. As in CGP,
    a resultant graph (c) and output program (d) are then extracted.}
  \label{fig:pcgp}
\end{figure*}

When $I_{start} = -1.0$, as in \autoref{fig:pcgp}, the input space is large and
nodes have a high probability of connecting directly to inputs. As this is not
desirable for complex programs, the $I_{start}$ parameter was tested in the
following experiments.

Due to the evolution of the positions, it is highly likely that no two nodes
occupy the same position, even between different genomes. Furthermore, over
evolution, nodes which are connected can have positional genes and connection
genes which are highly related. Finally, a node's connection positions depend
only on that node's position, which is in its genes, and not on the node's
placement in the genome or other nodes in the network. This allows node genes to
be exportable; the same genes in a different individual will form connections in
the same place. If multiple genes are exported together, entire sections of the
graph can be migrated between individuals. In PCGP, nodes can be added or
removed from a genome without disturbing the existing connection scheme, unlike
in CGP, where a node addition and deletion causes a shift in all downstream node
positions. This is the inspiration for the following study, where graph based
operators from other forms of GP are used in PCGP.

\section{Genetic operators}

GP has numerous genetic operators defined across its many implementations.
Genetic mutation and single point crossover have been used extensively, but
tree-based GP also mutates and crosses specific parts of a genome.
Autoconstructive evolution introduced many operators as part of a program
modification instruction set \cite{Spector2002}. Evolution of artificial neural
networks (ANN) \cite{Stanley2002} and gene regulatory networks (GRN)
\cite{Cussat2015} provide examples of genetic operators especially suited for
genomes that encode graphs, which is relevant to both CGP and PCGP. Parallel
distributed GP (PDGP) was inspired by ANNs and included a subgraph addition
mutation and a subgraph crossover method called subgraph active-active node
(SAAN) \cite{poli1997evolution}. A comparison of different crossover operators
and ideal parameters for these methods is presented in \cite{husa2018crossover}.
Here we define multiple genetic operators for both CGP and PCGP drawn from other
GP methods, ANN evolution, and GRN evolution. Operators are defined in
\textbf{bold} and common methods used by multiple operators are defined in
\textit{italics}.

\subsection{Mutation}

First, we define common methods used by multiple operators. We use the term
``computational nodes'' to refer to CGP or PCGP nodes which are neither inputs
nor outputs.

\textit{Node addition}: $m_{\delta}size_{min}$ computational nodes with random genes
are added to the end of the genome. In PCGP, these are then sorted into the
genome based on position.

\textit{Node deletion}: $m_{\delta}size_{min}$ computational nodes are randomly
selected from the genome and removed. In the event that there are fewer than
$m_{\delta}size_{min}$ computational nodes, all computational nodes are removed.

\textit{Subgraph addition}: $m_{\delta}size_{min}$ computational nodes are added
to the genome. The position, function, and parameter genes are randomly chosen,
but connection genes are selected randomly out from a pool. For each new node
$i$, this pool composed of all other new nodes with position $p < p_i$. An equal
number of randomly selected computational and input nodes with $p < p_i$ are
also added to the pool from the parent chromosome. By fixing the connection
genes, the new genetic material is guaranteed to either contain new subgraphs or
to create a subgraph with existing nodes. Due to the requirement of knowing
exact node positions for connection, this operation is available only in PCGP.

\textit{Subgraph deletion}: The parent genome is evaluated into functional
trees, both those which result in a final output (active) and those which don't
(inactive or junk). A tree with more than 1 computational node is chosen
randomly and up to $m_{delta}size{min}$ computational nodes are removed from it.

These methods are used in the following three mutation operators:

\textbf{(1) Gene mutation}: In CGP and PCGP, each gene of the computational
nodes has a $m_{node}$ chance of being replace by a new random value in $[0.0,
  1.0]$. Outputs are similarly mutated with a $m_{output}$ chance. In PCGP only,
the positions of the inputs are mutated with a $m_{input}$ chance. If
$m_{active}$ is true, this mutation is repeated until a gene in an active node
is mutated \cite{Goldman2013b}.

\textbf{(2) Mixed node mutate}: A random method between \textbf{gene mutation},
\textit{node addition}, and \textit{node deletion} is chosen according to
$m_{modify}$ and the size of the parent genome. A random number $r$ is chosen in
$[0.0, 1.0]$. If $r$ is less than $m_{modify}$, \textbf{gene mutation} is
selected. If $r$ is less than $m_{modify} + m_{add}$, \textit{node addition} is
selected. Otherwise, \textit{node deletion} is selected. $m_{add}$ is calculated
based on the number of nodes in the parent genome, $n$:

\begin{equation*}
  m_{add} = \frac{(n - size_{min})(1-m_{modify})}{size_{max} - size_{min}}
\end{equation*}





\textbf{(3) Mixed subgraph mutate}: Following the same logic as \textbf{mixed
  node mutate}, a method between \textbf{gene mutation}, \textit{subgraph
  addition}, and \textit{subgraph deletion} is chosen according to $m_{modify}$
and the size of the parent genome.


\subsection{Crossover}

For CGP, three crossover methods have been defined: \textbf{single point},
\textbf{random node}, and \textbf{proportional}. As PCGP allows for program
structure to be preserved during genetic transfer and contains additional
genetic material in the form of positions, further crossover methods can be
defined: \textbf{aligned node}, \textbf{output graph}, and \textbf{subgraph}.

\textbf{(1) Single point crossover}: In this classic crossover operator, a
single point in the two parent genomes is selected randomly. The genetic
material before this point is taken randomly from one parent and the genetic
material after this points is taken from the other parent. In CGP and PCGP, the
point is constrained to the beginning of a node's genetic material.

\textbf{(2) Random node crossover}: Nodes are randomly selected equally
from both parents. A child is constructed using randomly selected input and
output genes from both parents, the selected computational node genes from the first
parent, then finally the selected computational node genes from the second parent. The
ordering of the genetic material is important for CGP, but in PCGP the nodes and
their corresponding genes are ordered by their position.

\textbf{(3) Aligned node crossover}: This operator is only applicable
for PCGP. Nodes are first paired from each parent based on position proximity,
This operator then follows the same method as \textbf{random node crossover},
however nodes are randomly chosen from their position aligned pairs.

\textbf{(4) Proportional crossover}: This operator was previously
explored in \cite{Clegg2007}. The child's genetic material $C$, up to the
minimum size of both parents ($A$ and $B$), is combined using a vector of
randomly chosen weights, $w$:

\begin{equation}
  C_i = (1-w_i)A_i + w_iB_i~~\forall i
\end{equation}

If one parent genome was longer than the other, the remaining genetic material
is appended to the end of the child genome.

\textbf{(5) Output graph crossover}: Outputs from each parent are
randomly selected for the child genome. For each selected output, the full
functional graph resulting in this output is computed, and the set of all computational
nodes in the selected output graphs for each parent is used to construct the
child genome. Functional arity is ignored in this output trace, meaning that
inactive genetic material from 1-arity functions will be passed on to the child
genome. Otherwise, this operator only takes active nodes from each parent. If an
input is used in only one parent's selected output graph, it is passed on
to the child directly. Otherwise, each input is randomly selected from both
parents. As this operator assumes that the functional graph directly corresponds
to the transferred genetic material, it is only available in PCGP.

\textbf{(6) Subgraph crossover}: Similarly to \textbf{output graph
  crossover}, the functional graphs from the parent genomes are computed.
However, in this operator, active and inactive subgraphs from both parents are
randomly selected equally. Input and output genes are selected randomly from
both parents. As with \textbf{output graph crossover}, this operator is only
applicable to PCGP.

\section{Experiments}

\begin{table*}[h]
  \centering
  \begin{tabular}[t]{l|l|p{8cm}}
    Parameter & type & range\\
    \hline
    mutation & c & genetic, mixed node, mixed subgraph\\
    crossover & c & single point, proportional, random node, aligned node, output graph, subgraph\\
    $\lambda$ (population) & i & [1, 10]\\
    GA population & c & 20, 40, 60, 80, 100, 120, 140, 160, 200\\
    $I_{start}$ & r & [-1.0, -0.1]\\
    $r$ & r & [0.0, 1.0]\\
    $w$ & c & true, false\\
    $m_{active}$ & c & true, false\\
    $m_{input}$ & r & [0.0, 1.0]\\
    $m_{output}$ & r & [0.1, 1.0]\\
    $m_{node}$ & r & [0.1, 1.0]\\
    $m_{\delta}$ & r & [0.1, 0.5]\\
    $m_{modify}$ & r & [0.1, 0.9]\\
    $GA_{elitism}$ & r & [0.0, 0.8]\\
    $GA_{crossover}$ & r & [0.1, 1.0]\\
    $GA_{mutation}$ & r & [0.1, 1.0]
  \end{tabular}
  \caption{Ranges used in irace. The different range types are choice (c),
    integer (i), and real-valued (r). The precision for real-valued parameters
    was 0.1.}
  \label{tab:param_ranges}
\end{table*}

To explore the utility of these different genetic operators in CGP and PCGP, a
parameter study is done using irace \cite{Lopez-ibanez2016}. irace is an
automatic algorithm configuration package which selects from ranges of
parameters and explores the parameter space efficiently by focusing on high
performing parameter sets in a method known as racing.

The different genetic operators are parameterized and included with all CGP and
PCGP parameters for irace optimization. A $1+\lambda$ EA and a GA are used, and
the necessary parameters for the two are also optimized. The GA includes the
parameters $GA_{elitism}$, determining the number of top individuals retained
each generation, $GA_{crossover}$, the percentage of new individuals produced by
crossover, and $GA_{mutation}$, the percentage of individuals produced by
mutation. If these three sum to less than 1.0, random tournament winners are
added to the population unmodified. CGP and PCGP are evaluated separately, and
each is evaluated on an EA and GA separately, creating four different parameter
optimization cases.

These four cases are evaluated using nine benchmarks: three classification
problems, three regression problems, and three reinforcement learning or control
problems. The classification and regression problems are from the UCI machine
learning repository \footnote{\url{http://archive.ics.uci.edu/}}. The
classification problems are the breast cancer, diabetes, and glass datasets,
which are standard problems in classification and represent different challenges
in classification. The regression datasets are abalone, wine quality, and forest
fire area, which are also standard benchmark sets. The reinforcement learning
tasks are three locomotion tasks from the PyBullet library \cite{coumans2018}.
In these tasks, a robotic ant, cheetah, and humanoid must be controlled to walk
as far as possible from the starting point.

\begin{table*}[h]
  \begin{tabular}{l|*{3}{l}|*{3}{l}|*{3}{l}}
    & \multicolumn{3}{c|}{$e_0$} & \multicolumn{3}{c|}{$e_1$} & \multicolumn{3}{c}{$e_2$}\\
    & \multicolumn{3}{c|}{$1+\lambda$} & \multicolumn{3}{c|}{$1+\lambda$} & \multicolumn{3}{c}{GA}\\
    & \multicolumn{3}{c|}{CGP} & \multicolumn{3}{c|}{PCGP} & \multicolumn{3}{c}{CGP}\\
    class & C & R & RL & C & R & RL & C & R & RL\\
    mutation & node & gene & gene & gene & node & node & gene & gene & gene\\
    crossover &  &  &  &  &  &  & sp & sp & sp\\
    population & 4 & 9 & 4 & 6 & 3 & 8 & 20 & 120 & 200\\
    $I_{start}$ &  &  &  & -0.5 & -0.5 & -0.5 &  &  &\\
    $r$ & 0.2 & 0.1 & 0.6 & 0.5 & 0.4 & 0.1 & 0.4 & 0.1 & 0.8\\
    $w$ & 0 & 0 & 1 & 0 & 0 & 1 & 0 & 0 & 0\\
    $m_{active}$ & 1 & 1 & 0 & 0 & 0 & 0 & 0 & 1 & 1\\
    $m_{input}$ &  &  &  & 0.1 & 0 & 0.3 &  &  &\\
    $m_{output}$ & 0.2 & 0.1 & 0.3 & 0.3 & 0.2 & 0.8 & 0.1 & 0.5 & 0.3\\
    $m_{node}$ & 0.1 & 0.1 & 0.1 & 0.1 & 0.1 & 0.2 & 0.2 & 0.1 & 0.1\\
    $m_{\delta}$ & 0.4 &  &  &  & 0.2 & 0.2 &  &  &\\
    $m_{modify}$ & 0.6 &  &  &  & 0.9 & 0.5 &  &  & \\
    $GA_{elitism}$ &  &  &  &  &  &  & 0.1 & 0.2 & 0.1\\
    $GA_{crossover}$ &  &  &  &  &  &  & 0.2 & 0.2 & 0.2\\
    $GA_{mutation}$ &  &  &  &  &  &  & 0.3 & 0.6 & 0.2\\
    \hline
    & \multicolumn{3}{c|}{$e_3$} & \multicolumn{3}{c|}{$e_4$} &\multicolumn{3}{c}{$e_5$}\\
    & \multicolumn{3}{c|}{GA} & \multicolumn{3}{c|}{$1+\lambda$} &\multicolumn{3}{c}{GA}\\
    & \multicolumn{3}{c|}{PCGP} & \multicolumn{3}{c|}{CGP} &\multicolumn{3}{c}{CGP}\\
    class & C & R & RL & \multicolumn{3}{c|}{all} &\multicolumn{3}{c}{all}\\
mutation& gene & gene & node& & gene & & & gene&\\
crossover& prop & prop & output&  & & & & prop&\\
population& 120 & 80 & 200& & 5 & & & 50&\\
$I_{start}$& -0.9 & -0.2 & -0.3& & & & & &\\
$r$& 0.5 & 0.1 & 0.2 & & 0.0 & & & 0.0 & \\
$w$& 0 & 0 & 1& & 0 & & & 0 &\\
$m_{active}$& 0 & 1 & 1 & & 0 & & & 0 &\\
$m_{input}$& 0 & 0 & 0.2 & & & & & &\\
$m_{output}$& 0.2 & 0.4 & 0.3 & & 0.1 & & & 0.2 &\\
$m_{node}$& 0.1 & 0.1 & 0.4 & & 0.1 & & & 0.2 &\\
$m_{delta}$&  &  & 0.3 & & & & & &\\
$m_{modify}$&  &  & 0.4 & & & & & &\\
$GA_{elitism}$& 0.2 & 0.4 & 0.4 & & & & & 0.04 &\\
$GA_{crossover}$& 0.1 & 0.2 & 0.5 & & & & & 0.5 &\\
$GA_{mutation}$& 0.7 & 0.4 & 0.1 & & & & & 1.0 &\\
\\
  \end{tabular}
  \caption{The optimized parameter sets, $e_0$ through $e_3$, found by irace,
    and the default parameter sets $e_4$ and $e_5$. The different problem types
    are classification (C), regression (R), and reinforcement learning (RL). The
    mutation methods used are genetic mutation (gene) and mixed node mutation
    (node). The crossover methods used are single point (sp), proportional
    (prop), and output graph (output).}
  \label{tab:comparison_params}
\end{table*}

\begin{figure*}[h!]
    \centering
    \includegraphics[width=0.49\textwidth]{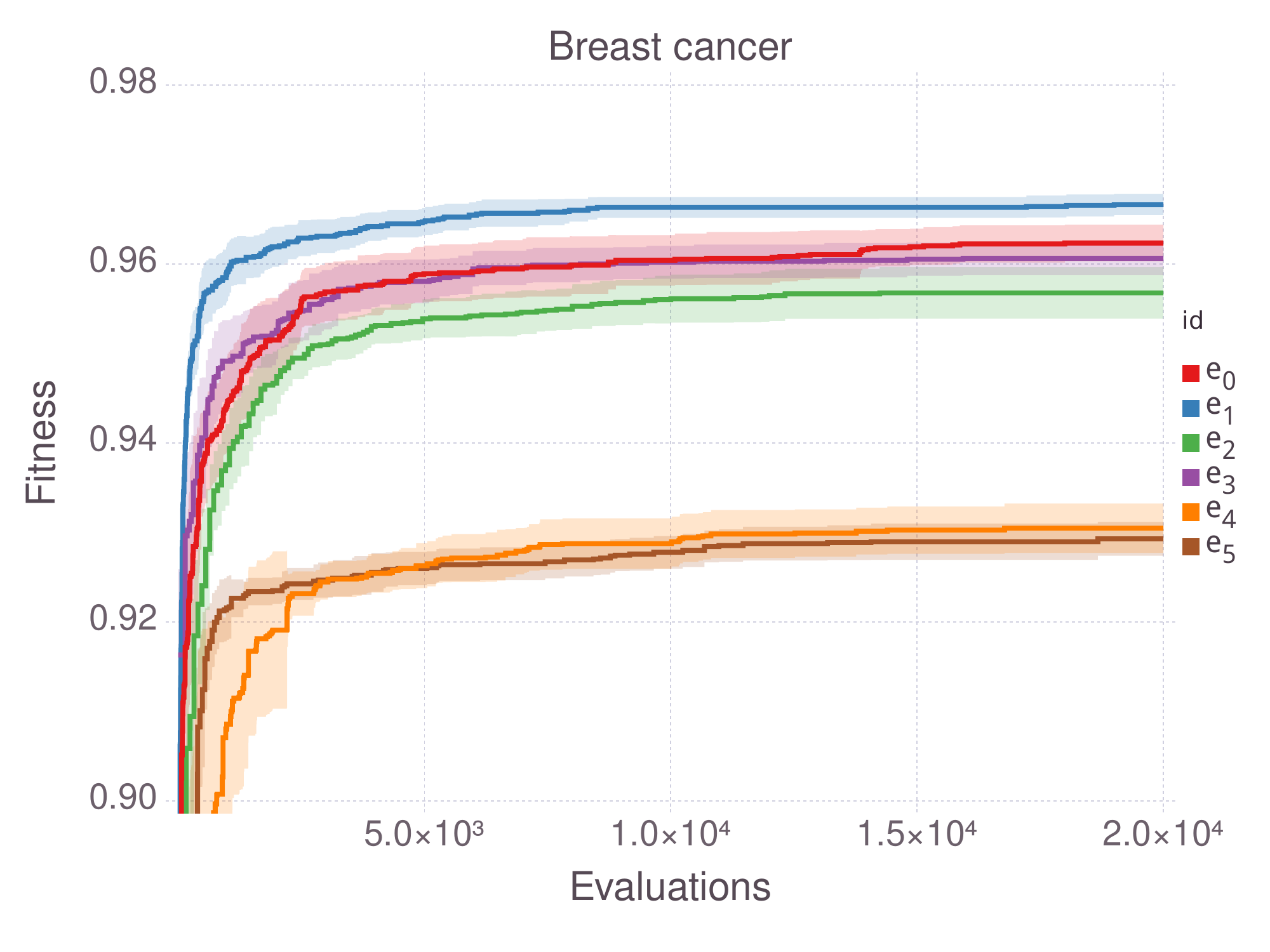}
    \includegraphics[width=0.49\textwidth]{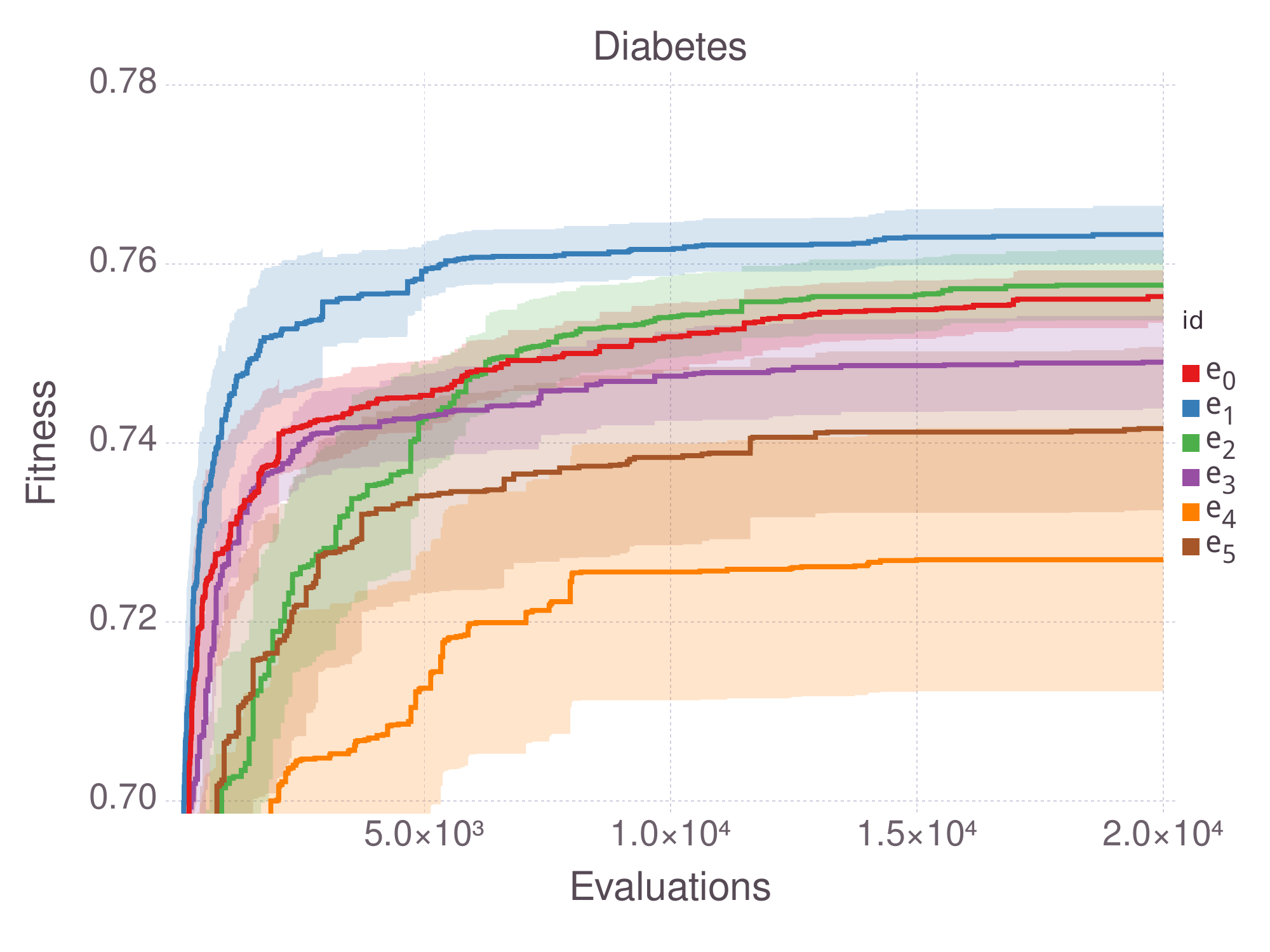}
    \includegraphics[width=0.49\textwidth]{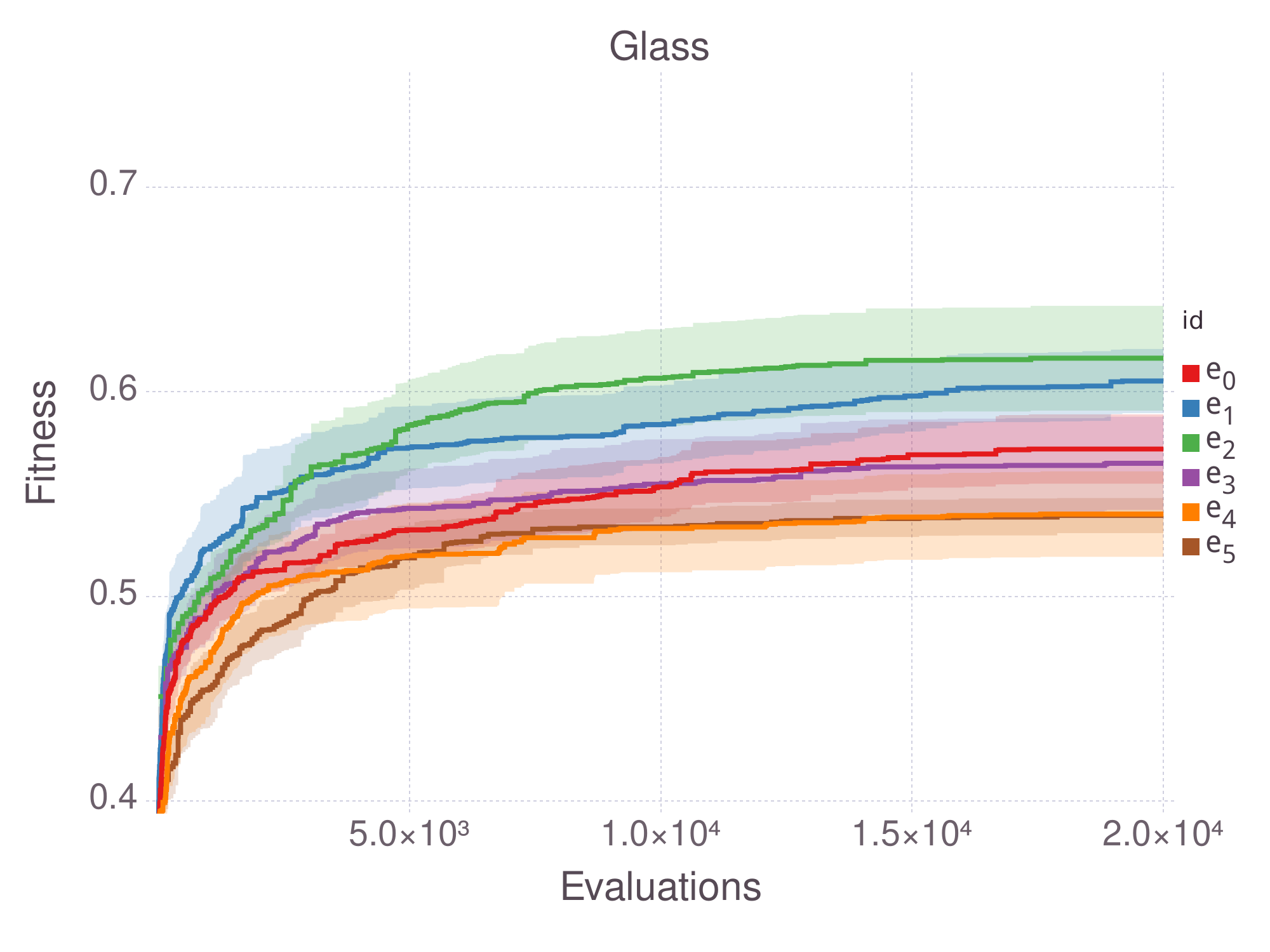}
    \includegraphics[width=0.49\textwidth]{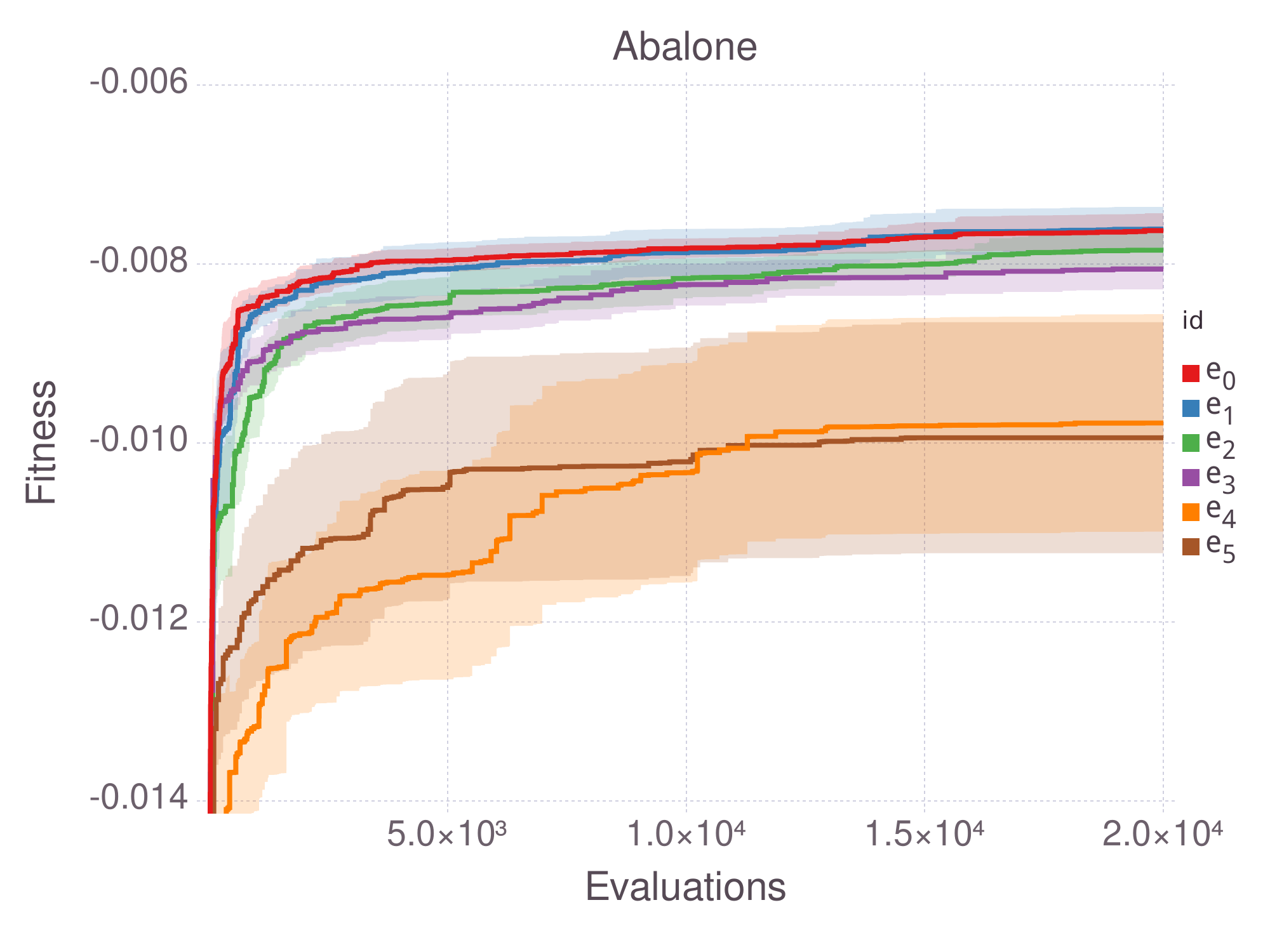}
    \includegraphics[width=0.49\textwidth]{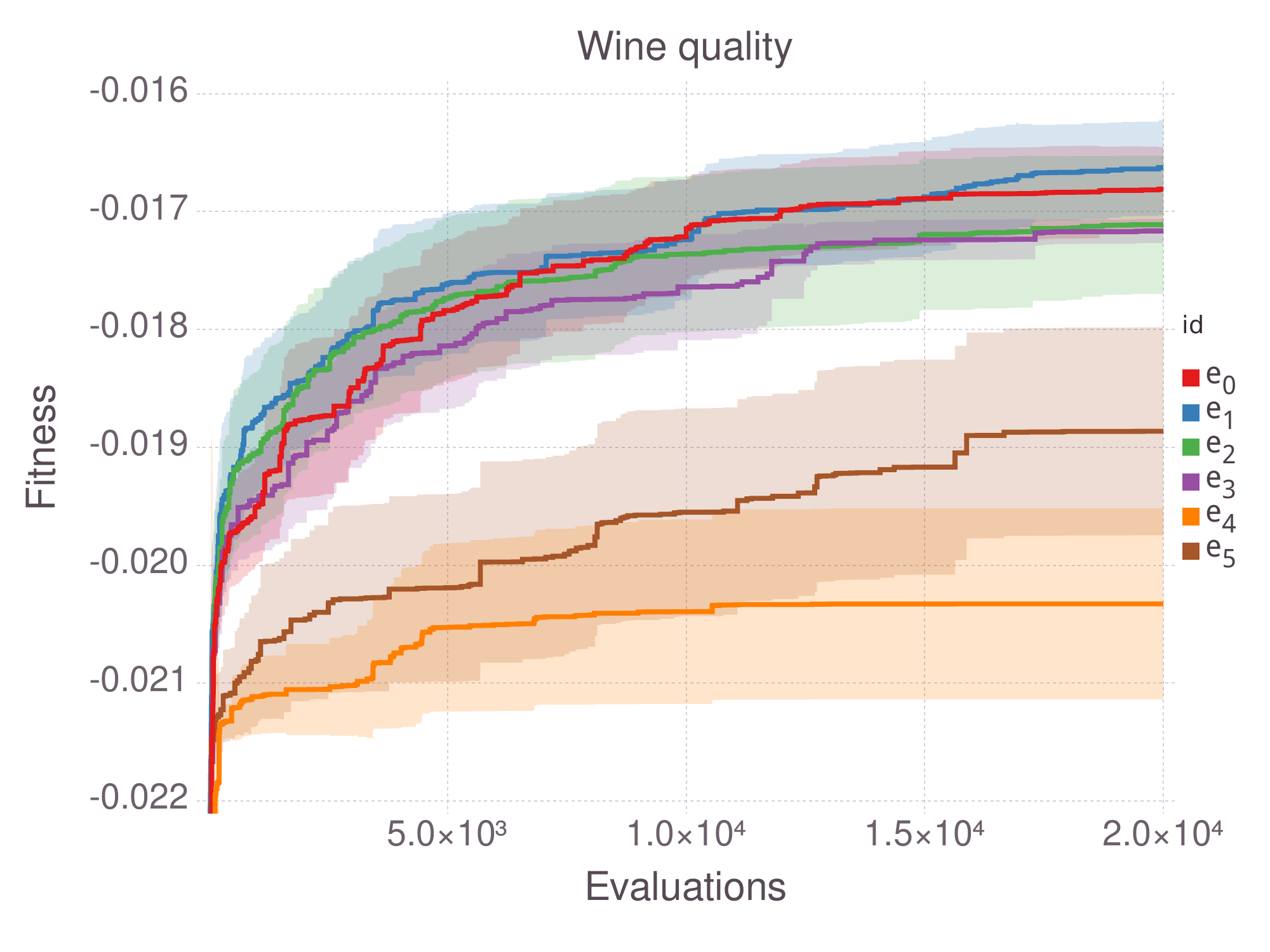}
    \includegraphics[width=0.49\textwidth]{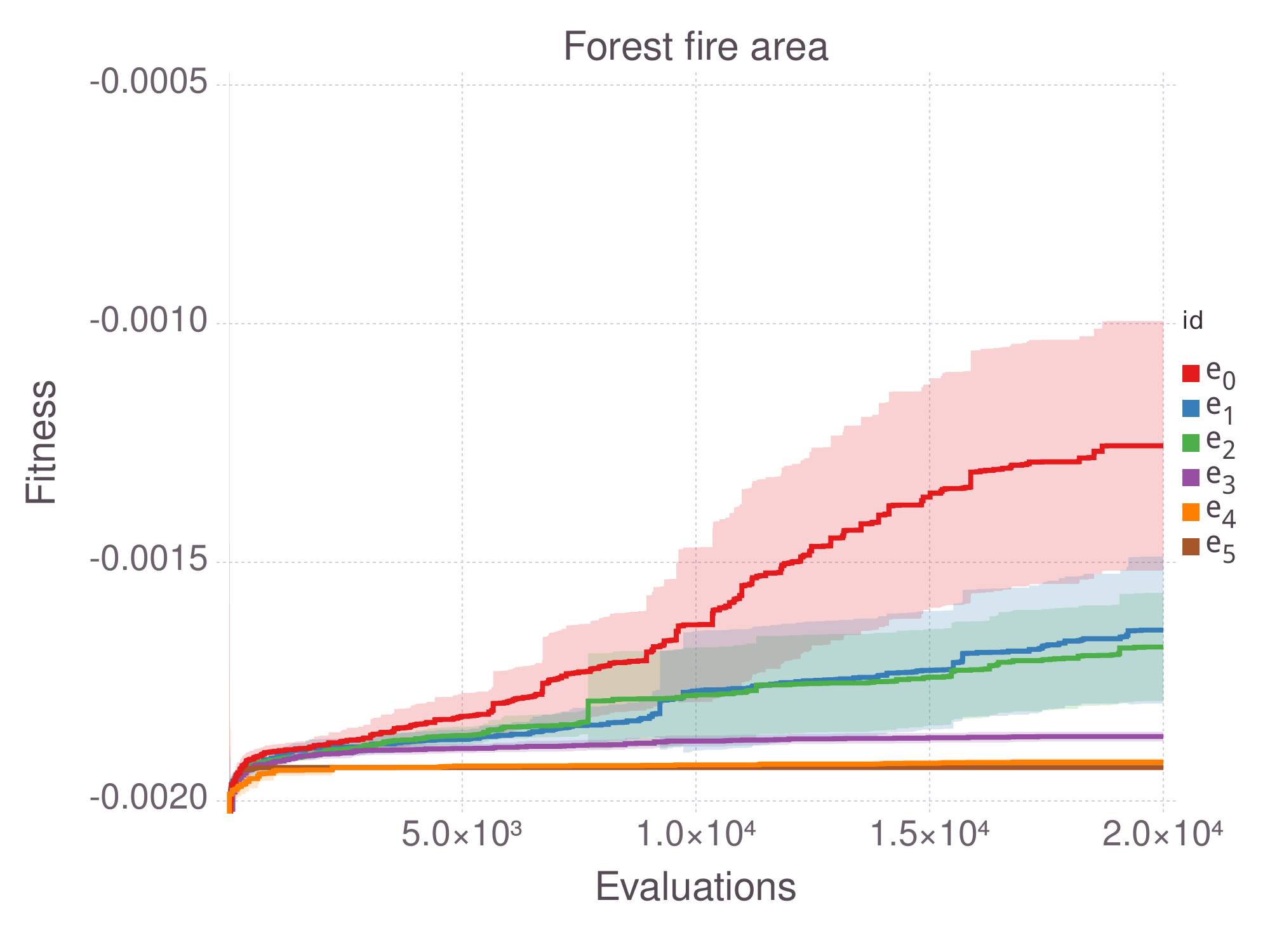}
    \includegraphics[width=0.49\textwidth]{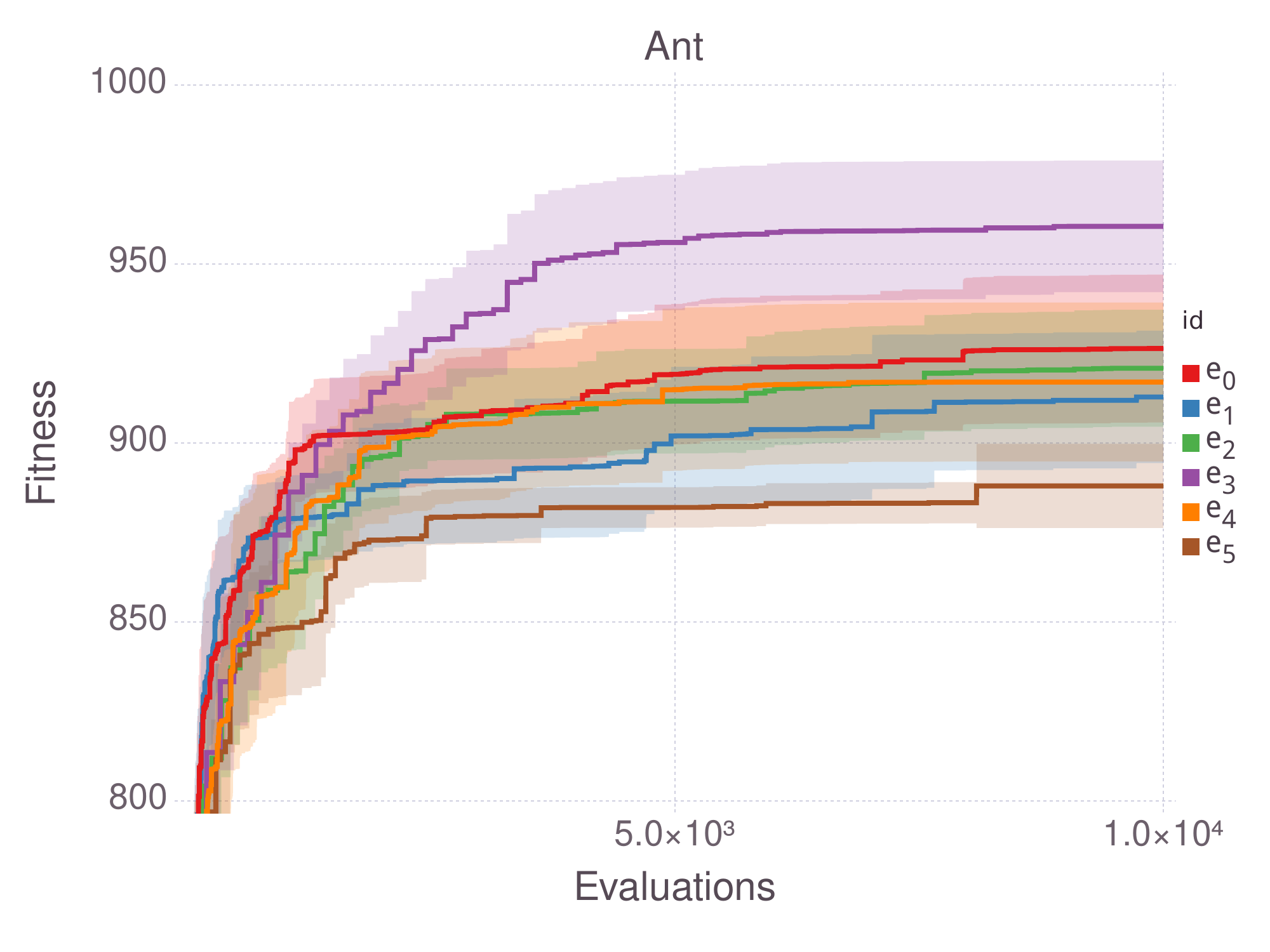}
    \includegraphics[width=0.49\textwidth]{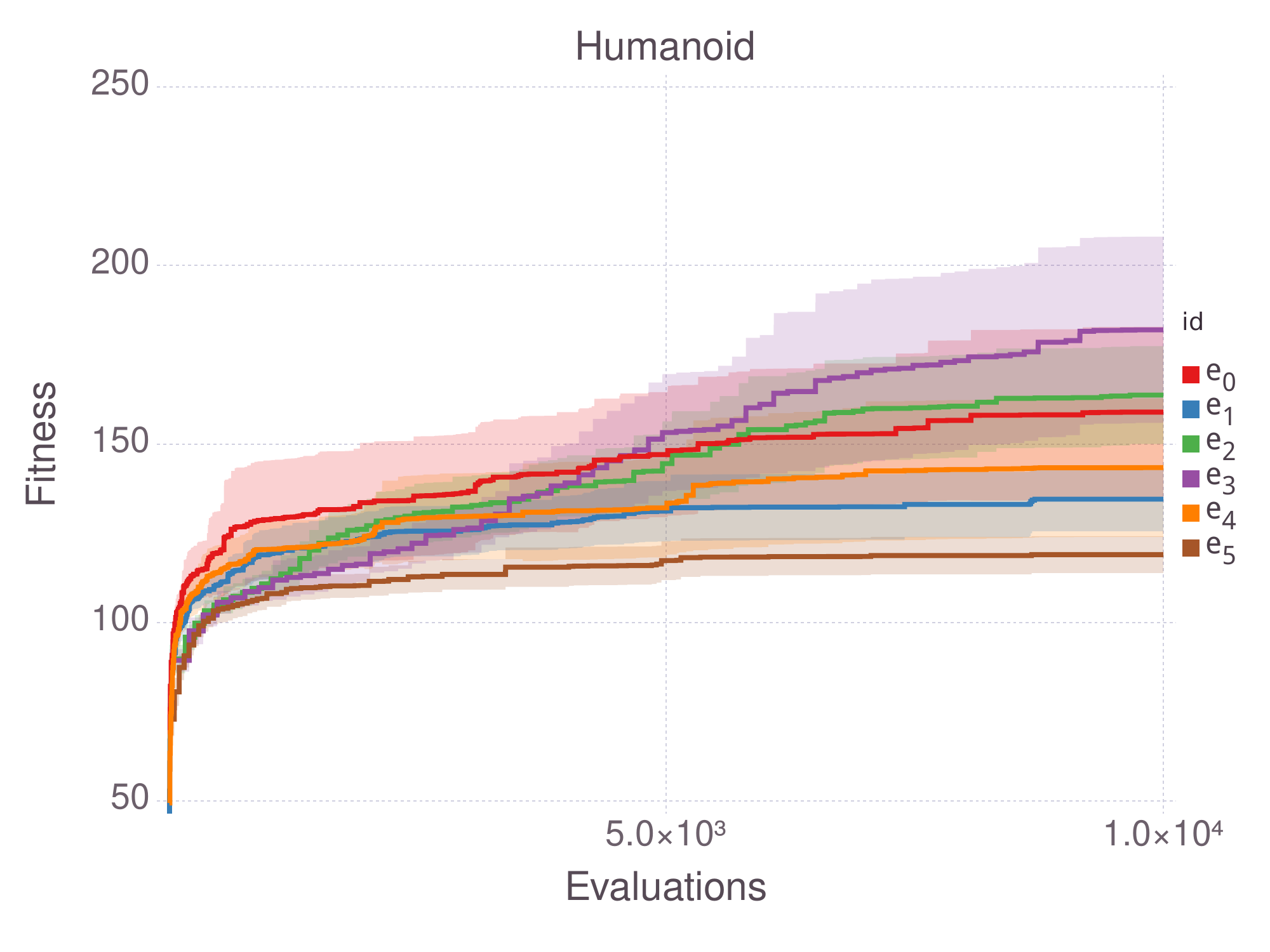}
    \caption{Comparison of the different methods on eight of the nine problems.
      Cheetah locomotion is omitted for space, due to similar results for all
      tested methods. Lines show the average of the best individuals and ribbons
      show one standard deviation from 20 trials.}
    \label{fig:comparison_results}
\end{figure*}

\section{Method comparison}

First, we compare the four different methods, $1+\lambda$ EA and a GA, using
both CGP and PCGP, using the best parameters found by irace for the set of three
problems of each type. The optimized methods are compared to the default
parameters of CGP, $e_4$ and the parameters reported in \cite{Clegg2007}, $e_5$,
as it is a similar work which uses crossover and a floating point
representation. The crossover rate was chosen based on the results from
\cite{Clegg2007}, although that work includes an interesting study of a variable
crossover rate, which was not implemented for these experiments. The optimized
and default parameters are presented in \autoref{tab:comparison_params}.

These parameters were used in 20 evolutions on each of the nine problems. To
compare the different population sizes and methods, the results are compared
based on the number of fitness evaluations, not by generation. For the
classification and regression problems, each evolution was run for 20000
evaluations and 10000 for the RL problems, as these problems were far more
computationally expensive. The computational budget in terms of nodes was also
considered; as some methods have a variable genome size, allowing them to
increase their computational limit in a way static methods cannot, the variable
methods were constrained to $[0.5, 1.5]$ of the genome size for the static
methods.

Overall, the results, displayed in \autoref{fig:comparison_results}, show that
the optimized parameter sets perform better than the two default parameter sets,
$e_4$ and $e_5$. Between the four optimized methods, there is no clear winner.
$e_1$, the $1+\lambda$ PCGP EA, is superior on two of the classification
problems, but not on the other problems, especially RL,
where it is the worst of the optimized methods. $e_3$, the PCGP GA, is superior
on two of the RL problems, but performs badly on
classification and regression.

It is worth noting that, while the GA methods $e_2$ and $e_3$ don't always
perform the best, their results are competitive with the $1+\lambda$ EA methods.
While this comparison is based on number of fitness evaluations, the GA has the
advantage of massive parallelization that the EA doesn't. In the GA, a
generation of up to 200 individuals can be evaluated in parallel, giving similar
results to the EA methods in a fraction of the time.

The main conclusion that can be drawn from this comparison is that the choice of
method, and the parameters of the chosen method, can greatly improve CGP
performance. To better understand appropriate parameters, therefore, we next
analyze the full results from irace, beyond the best individual used in these
comparisons.

\begin{figure*}[h]
  \includegraphics[width=1.0\textwidth]{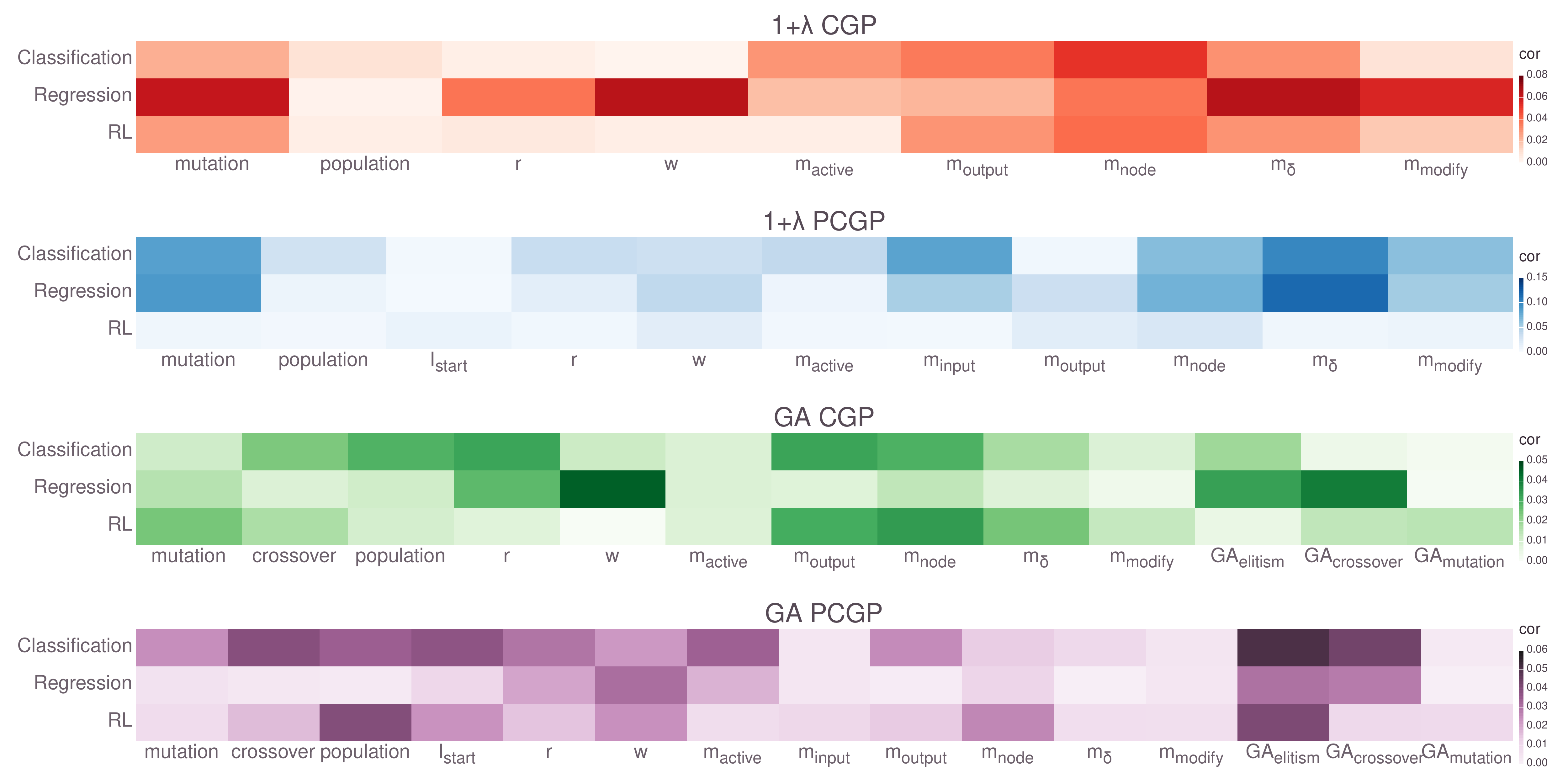}
  \caption{Correlation of all parameters with evolutionary fitness, using all
    parameter sets explored by irace. Each color represents a different method,
    and the color intensity represents the correlation with fitness for that
    parameter. Correlations are divided by problem class (classification,
    regression, and RL), using the final fitness results from all problems in
    the specific class}.
  \label{fig:correlations}
\end{figure*}

\section{Parameter study}

\begin{figure*}[h]
    \centerline{\includegraphics[width=1.0\textwidth]{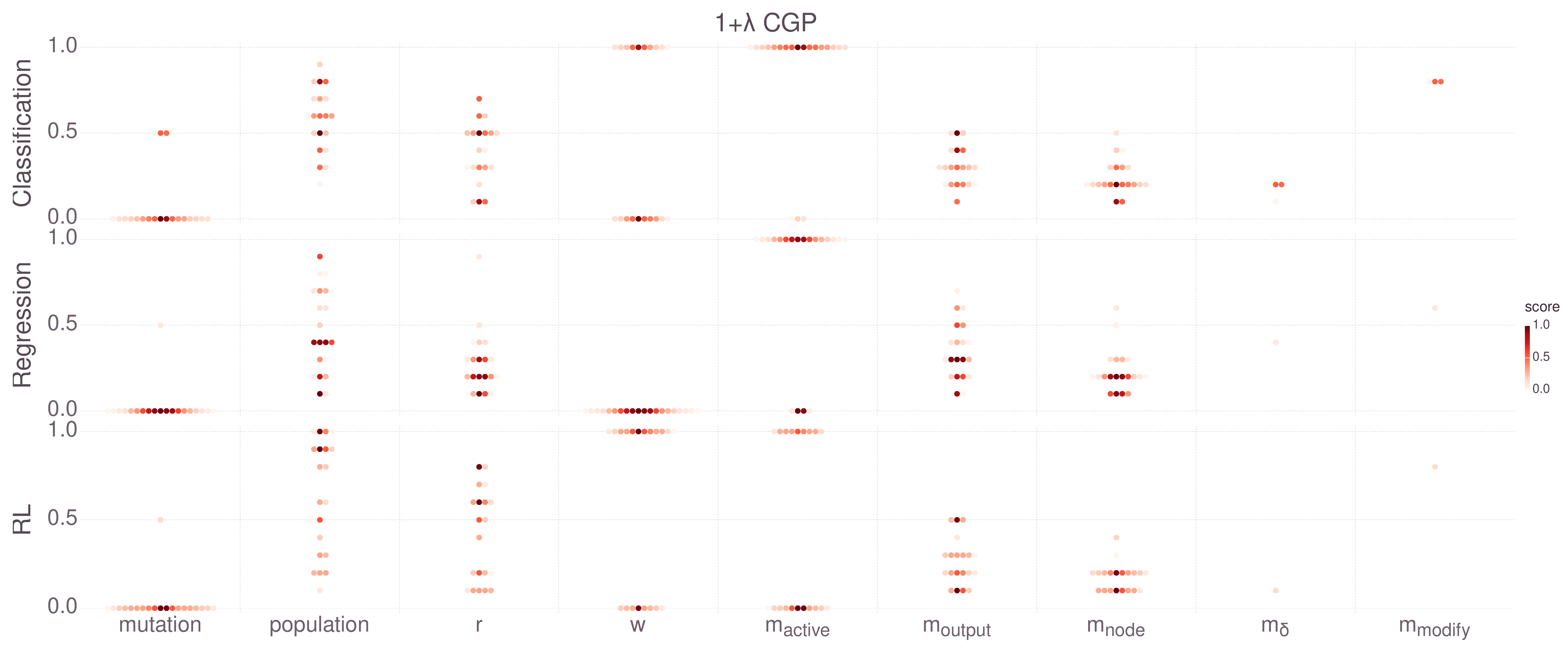}}
    \vspace{10mm}
    \centerline{\includegraphics[width=1.0\textwidth]{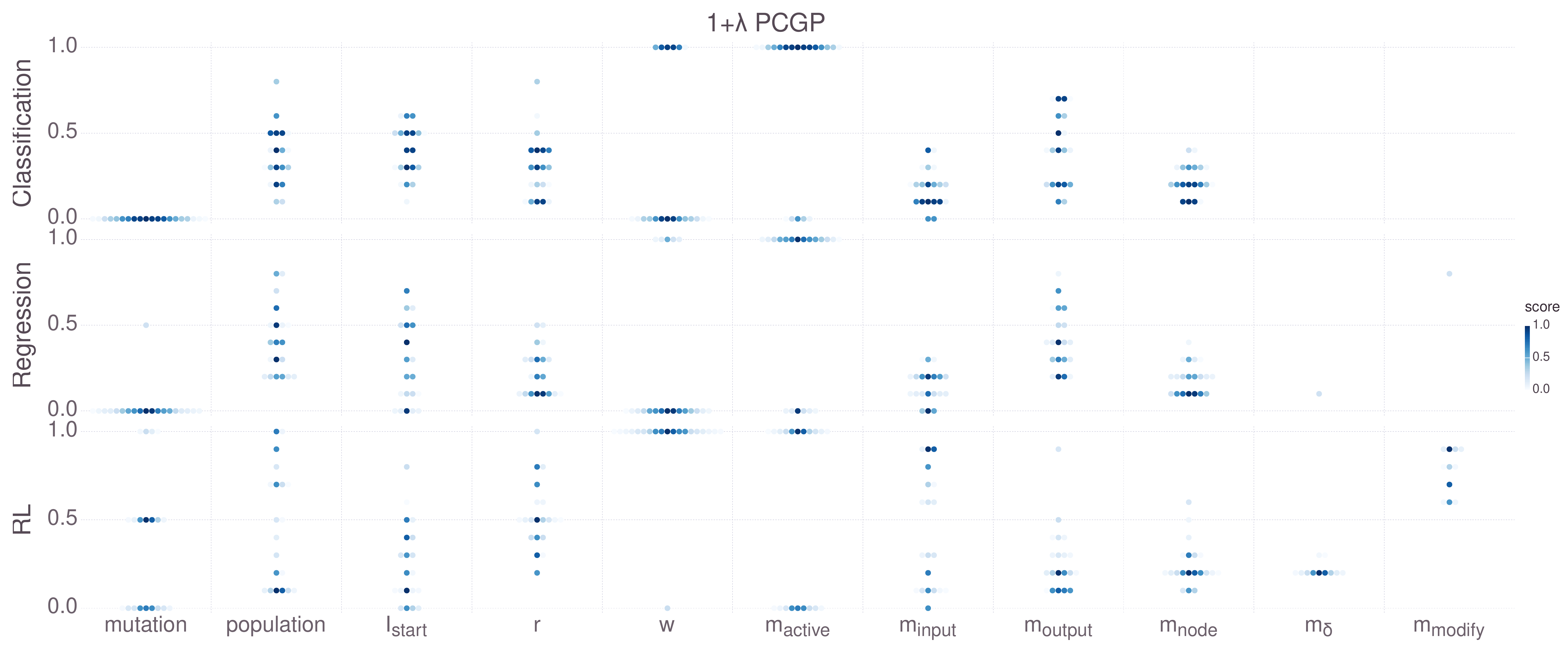}}
    \caption{Parameters for the top 20 $1+\lambda$ EA configurations, for CGP and
      PCGP}
    \label{fig:oneplus_params}
\end{figure*}

\begin{figure*}[h]
    \centerline{\includegraphics[width=1.0\textwidth]{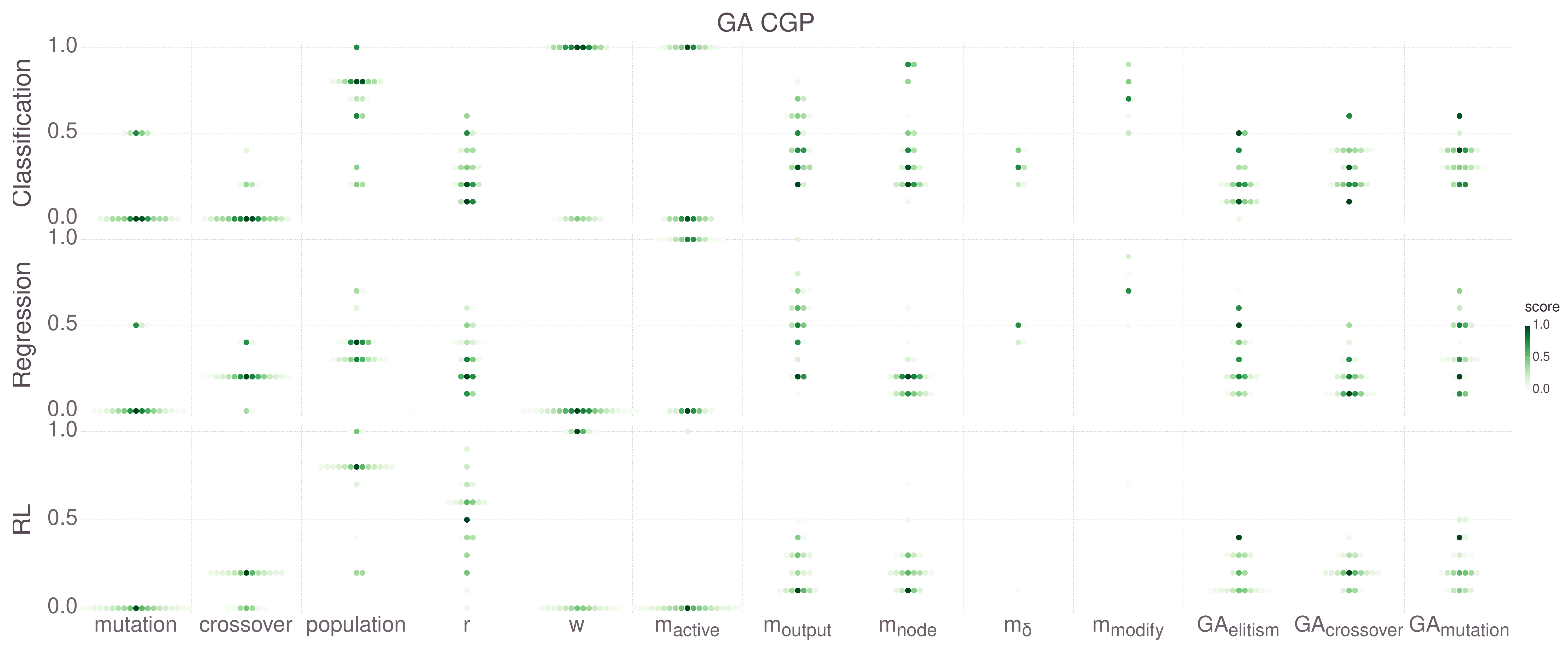}}
    \vspace{10mm}
    \centerline{\includegraphics[width=1.0\textwidth]{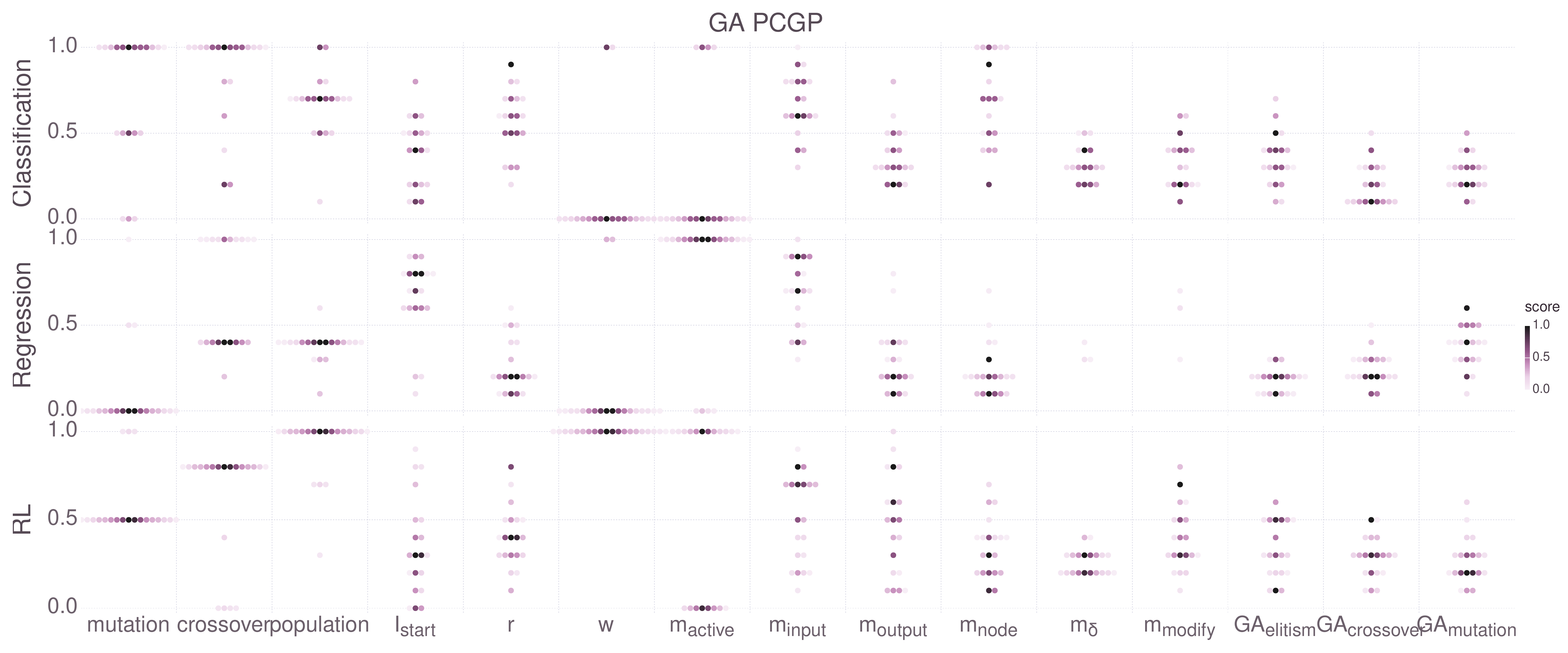}}
    \caption{Parameters for the top 20 GA configurations, for CGP and PCGP.}
    \label{fig:GA_params}
\end{figure*}

Finally, we explore the parameter choices produced by irace. The top 20
parameters, called expert parameter sets, for each method are displayed for the
different class types in \autoref{fig:oneplus_params} and
\autoref{fig:GA_params}, and the correlation between each parameter and
evolutionary fitness is shown in \autoref{fig:correlations}. In
\autoref{fig:oneplus_params} and \autoref{fig:GA_params}, all parameters are
represented as values between 0.0 and 1.0. To achieve this, the mutation
operators were ordered as [genetic, mixed node, mixed subtree], being [0.0, 0.5,
  1.0], and the crossover operators were ordered [single point, proportional,
  random node, aligned node, output graph, subgraph], being [0.0, 0.2, 0.4, 0.6,
  0.8, 1.0]. The population parameter represents $\frac{\lambda}{10}$ for the
$1+\lambda$ EA and $\frac{population}{200}$ for the GA. Finally, $I_{start}$ is
represented as $1+I_{start}$.

For CGP with the $1+\lambda$ EA, there is a clear preference for gene mutation
over mixed node mutation across all problems. Mutation rates are important, and
most expert parameters have a slightly higher output mutation rate than
node mutation rate, with $m_{output}$ reaching as high as 0.5 in many experts.
Node weights, $w$, especially have an impact for regression, where they
should not be used. Active mutation, $m_{active}$, appears useful mostly for
classification and regression and isn't correlated with fitness for RL tasks.
Finally, $\lambda$ appears to have little effect on the outcome, and $r$ appears
to impact the result only in the case of regression tasks, when it should be
low.

The expert parameter results for PCGP with the $1+\lambda$ EA are similar to
that with CGP. Gene mutation is a clear winner, using a higher mutation rate for
outputs than inputs and nodes. $\lambda$, $I_{start}$, and $r$ aren't strongly
correlated with fitness. Node weights appear useful in RL problems but are
detrimental in classification and regression problems, and active mutation
appears generally useful. An interesting difference in $1+\lambda$ EA PCGP is
the prevalence of mixed node mutation in expert parameter sets for RL problems.
However, due to a high $m_{modify}$ of these parameter sets, the main functional
mutation in these expert sets remained a gene modification mutation, with rare
node addition and deletion events.

For CGP, the expert parameters for a GA are very similar to the $1+\lambda$ CGP
EA expert parameters. Population is much more important than in the $1+\lambda$
EA, with medium to large populations (100 to 200 individuals) showing an
advantage in classification and regression problems. Genetic mutation is the
clear choice for a mutation operator, while crossover is split between single
point for classification and proportional for regression and RL. Elitism is
rather high, reaching 50\% in some expert sets. Crossover rates are low, except
in the case of classification, where it has little bearing on the final outcome.

The parameter results for PCGP when using a GA are very interesting and
different from all other sets. Here, we see usage of the other mutation and
crossover operators; classification prefers mixed subtree mutation and subgraph
crossover, regression uses gene mutation and random node crossover, and RL uses
mixed node mutation with output graph crossover. The population is somewhat
problem dependent, but is especially important in RL, where large populations
are favorable. Node weights are highly preferred in RL, but not in regression or
classification. Elitism has a large impact on the final fitness, although the
values for RL and classification are spread almost evenly between 0.1 and 0.5.

Considering the success of $e_3$ on the RL problems, the PCGP GA parameters show
that output graph crossover and node-based mutation can be viable strategies for
PCGP evolution. It's notable also that subgraph crossover was used in expert
regression sets and favored in classification, showing that graph-based
operations can be useful generally. The RL problems have outputs corresponding
to the control of different limbs, which may offer more modularity than the
different classes of classification.

\section{Conclusion}

Positional CGP opens the possibility of doing graph operations during CGP
evolution. The experiments in this work demonstrate that there is the potential
for improvement of CGP's evolution, even if no single method proposed is
universally dominant. The possibilities in improving CGP evolution are expanded
by PCGP, and more work is needed to explore these potential improvements.

Some of the parameters explored in this work are at the level of evolution and
require global coordination. Others, such as $r$, $I_{start}$, $w$, etc, could
be included at the level of the genome. Even the choice of CGP or PCGP could be
a binary parameter within the genome, deciding if the positional genes are used
or not. This would allow an individual optimization of the hyper-parameters and
reduce the burden of parameter choice.

The global parameters could also benefit from dynamic change over evolution. In
\cite{Clegg2007}, a variable crossover rate which begins high and reduces to 0
as the population converges is used. Adaptive mutation rates have also been
proven to increase search for the $1+(\lambda, \lambda)$ EA \cite{Doerr2018} and
could benefit CGP.

Other methods of evolution are also made possible by this work. CMA-ES could
easily be used to evolve floating point CGP or PCGP. In \cite{Harding2013}, an
island-based $1+\lambda$ EA is used. The graph-based crossover methods presented
in this work might the integration of experts in that scheme. In
\cite{zaefferer2018linear}, multiple distance metrics for GP are evaluated.
These could be used in CGP or PCGP to introduce speciation to the GA, achieving
a similar effect to the island-based model.

Finally, this work offers a guide to CGP configuration, including parameters for
a successful GA evolution, which has not been the standard for CGP. Given the
quality of the results attained by CGP, the benefits of parallelism could be
used to solve many problems to which GP has yet to be applied.

\subsection*{Acknowledgments}
This work is supported by ANR-11-LABX-0040-CIMI, within programme ANR-11-IDEX-0002-02.

\bibliographystyle{apalike}
\bibliography{main}

\end{document}